\documentclass[11pt]{article}
\usepackage[margin=1in]{geometry}
\usepackage{lmodern}
\usepackage[round]{natbib}

\usepackage[T1]{fontenc}
\usepackage{amsmath,amssymb,bm}
\usepackage{graphicx}
\usepackage{booktabs,multirow,makecell,colortbl}
\usepackage[table]{xcolor}
\usepackage{float}
\usepackage{algorithm}
\usepackage{algpseudocode}
\usepackage{microtype}
\usepackage{enumitem}
\usepackage{caption}
\usepackage[colorlinks=true,linkcolor=black,citecolor={blue!50!black},urlcolor={blue!50!black}]{hyperref}
\usepackage{url}
\usepackage[capitalize,noabbrev]{cleveref}

\newif\ifhavesecondary
\havesecondarytrue
\newif\ifhaveround
\haveroundtrue
\newcommand{\nCAFull}{85.67}
\newcommand{\nCANoCL}{78.46}
\newcommand{\nCANoCR}{86.50}
\newcommand{\nCANoHNM}{85.16}
\newcommand{\nCANoSil}{84.99}
\newcommand{\nCARec}{85.33}
\newcommand{\nCDNoCL}{+7.21}
\newcommand{\nCDNoCLAbs}{7.21}
\newcommand{\nCDNoCR}{\textminus{}0.83}

\newcommand{\nCDNoHNM}{+0.51}
\newcommand{\nCDNoHNMAbs}{0.51}
\newcommand{\nCDNoSil}{+0.68}
\newcommand{\nCDNoSilAbs}{0.68}
\newcommand{\nCDRec}{+0.34}

\newcommand{\nCcsslBMean}{89.01}
\newcommand{\nCcsslBSd}{2.37}
\newcommand{\nCcsslBa}{90.47}
\newcommand{\nCcsslBb}{85.67}
\newcommand{\nCcsslBc}{90.90}
\newcommand{\nCcsslRDiff}{+0.03}

\newcommand{\nCcsslRMean}{89.05}
\newcommand{\nCcsslRSd}{2.40}
\newcommand{\nCcsslRa}{90.55}
\newcommand{\nCcsslRb}{85.66}
\newcommand{\nCcsslRc}{90.93}

\newcommand{\nCgainDAFAbs}{7.01}

\newcommand{\nCgainRecB}{+0.08}

\newcommand{\nCrecBMean}{88.94}
\newcommand{\nCrecBSd}{2.63}
\newcommand{\nCrecBa}{89.96}
\newcommand{\nCrecBb}{85.33}
\newcommand{\nCrecBc}{91.52}
\newcommand{\nCrecRDiff}{+0.02}

\newcommand{\nCrecRMean}{88.96}
\newcommand{\nCrecRSd}{2.65}
\newcommand{\nCrecRa}{90.02}
\newcommand{\nCrecRb}{85.32}
\newcommand{\nCrecRc}{91.54}

\newcommand{\nCrecVsDAFBAbs}{6.94}
\newcommand{\nCroundMaxDiff}{0.03}
\newcommand{\nCsupBMean}{85.87}
\newcommand{\nCsupBSd}{4.92}
\newcommand{\nCsupBa}{90.55}
\newcommand{\nCsupBb}{79.07}
\newcommand{\nCsupBc}{87.98}
\newcommand{\nCsupRDiff}{+0.02}

\newcommand{\nCsupRMean}{85.88}
\newcommand{\nCsupRSd}{4.97}
\newcommand{\nCsupRa}{90.60}
\newcommand{\nCsupRb}{79.02}
\newcommand{\nCsupRc}{88.03}
\newcommand{\nCwinsRecB}{2}
\newcommand{\nDAF}{82.0}
\newcommand{\nDtDadr}{77}

\newcommand{\nDtDualstream}{78}
\newcommand{\nDtSdnetMulti}{80}
\newcommand{\nDtSdnetSingle}{77}
\newcommand{\nDtTranslation}{75}
\newcommand{\nDtUNetMulti}{79}
\newcommand{\nDtUNetSingle}{76}
\newcommand{\nMAFull}{80.66}
\newcommand{\nMANoBnd}{77.99}
\newcommand{\nMANoCL}{78.51}
\newcommand{\nMANoCR}{80.42}
\newcommand{\nMANoFeat}{79.11}
\newcommand{\nMANoHNM}{80.07}
\newcommand{\nMANoMem}{81.14}
\newcommand{\nMANoSil}{80.54}
\newcommand{\nMARec}{77.69}
\newcommand{\nMDNoBnd}{+2.67}
\newcommand{\nMDNoBndAbs}{2.67}
\newcommand{\nMDNoCL}{+2.15}
\newcommand{\nMDNoCLAbs}{2.15}
\newcommand{\nMDNoCR}{+0.24}
\newcommand{\nMDNoCRAbs}{0.24}
\newcommand{\nMDNoFeat}{+1.55}

\newcommand{\nMDNoHNM}{+0.59}

\newcommand{\nMDNoMem}{\textminus{}0.48}
\newcommand{\nMDNoMemAbs}{0.48}
\newcommand{\nMDNoSil}{+0.12}
\newcommand{\nMDNoSilAbs}{0.12}
\newcommand{\nMDRec}{+2.97}

\newcommand{\nMcsslBacc}{87.27}
\newcommand{\nMcsslFgbg}{87.73}
\newcommand{\nMcsslMean}{80.87}
\newcommand{\nMcsslPooled}{81.00}
\newcommand{\nMcsslSd}{0.17}
\newcommand{\nMcsslsa}{80.66}
\newcommand{\nMcsslsb}{81.07}
\newcommand{\nMcsslsc}{80.88}
\newcommand{\nMgainRec}{+2.14}
\newcommand{\nMgainRecAbs}{2.14}
\newcommand{\nMgainReca}{+2.97}

\newcommand{\nMgainRecb}{+2.57}

\newcommand{\nMgainRecc}{+0.89}

\newcommand{\nMgainSLAbs}{1.77}

\newcommand{\nMgainSupHundredAbs}{3.23}
\newcommand{\nMrecMean}{78.73}
\newcommand{\nMrecSd}{0.95}

\newcommand{\nMrecsa}{77.69}
\newcommand{\nMrecsb}{78.50}
\newcommand{\nMrecsc}{79.99}
\newcommand{\nMsupFifty}{76.01}
\newcommand{\nMsupFiftyBacc}{85.23}
\newcommand{\nMsupFiftyFgbg}{84.41}
\newcommand{\nMsupGapSL}{+0.14}

\newcommand{\nMsupHGapSL}{\textminus{}1.66}

\newcommand{\nMsupHundred}{77.64}
\newcommand{\nMsupHundredBacc}{87.49}
\newcommand{\nMsupHundredFgbg}{84.62}
\newcommand{\nMwinsRec}{3}
\newcommand{\nSL}{79.10}
\newcommand{\nSLsupFifty}{75.87}

\newcommand{\nSLtEnsemble}{76.87}
\newcommand{\nSLtMcDropout}{77.98}

\newcommand{\nSLtSoftmax}{76.18}

\definecolor{csslblue}{HTML}{2A78D6}
\definecolor{recipeorange}{HTML}{EB6834}
\definecolor{ink2}{HTML}{52514E}
\definecolor{rowhl}{HTML}{EAF2FC}
\newcommand{\ours}[1]{\textbf{#1}}
\newcommand{\Lsup}{\mathcal{L}_{\mathrm{sup}}}

\newcommand{\Lcr}{\mathcal{L}_{\mathrm{CR}}}
\newcommand{\Lcl}{\mathcal{L}_{\mathrm{CL}}}
\newcommand{\CE}{\mathrm{CE}}
\newcommand{\Dice}{\mathrm{Dice}}
\newcommand{\sg}{\operatorname{sg}}
\newcommand{\nrm}{\operatorname{norm}}
\let\origpm\pm
\renewcommand{\pm}{\ensuremath{\,{\scriptstyle\origpm}\,}}
\setlist{itemsep=2pt,topsep=3pt}
\title{Contrastive Semi-Supervised Segmentation with Class-wise Patches and Silhouette-Adaptive Consistency}

\author{Prashant Pandey\\
Indian Institute of Technology Delhi\\
Contact: \href{mailto:getprashant57@gmail.com}{getprashant57@gmail.com} \\
Code: \url{https://github.com/prinshul/CSSL}}
\date{}

\begin{document}
\maketitle

\begin{abstract}
We present CSSL, a contrastive semi-supervised method for medical image segmentation. CSSL samples class-wise patches from labelled images and from confident pseudo-labels, separates their embeddings with a supervised contrastive loss with mixed hard negatives, and uses the Silhouette coefficient of the embedding clusters to set the strength of a feature-perturbation consistency loss. On MoNuSeg, patch embeddings are obtained by masked pooling of the encoder features of the unmasked image, nucleus rims form an additional contrastive class and a memory bank stores past embeddings; on CHAOS, patches are pixel-masked and no memory bank is used. The CSSL losses are added to a weak-to-strong pseudo-labelling recipe. We follow the protocols of published baselines: Self-Loop on MoNuSeg and DAFNet on CHAOS. On MoNuSeg with 50\% labels, CSSL reaches \nMcsslMean\pm\nMcsslSd{} nuclei F1 over three label splits, compared with \nSL{} for Self-Loop and \nMrecMean{} for the same recipe without CSSL. On CHAOS T2 with 25\% labels, CSSL reaches \nCcsslBMean{} Dice, compared with \nDAF{} for DAFNet and \nCrecBMean{} for the recipe alone. Ablations show that the contrastive branch and, on MoNuSeg, the boundary class and feature-level pooling account for most of the improvement.
\end{abstract}

\section{Introduction}
\label{sec:intro}

Pixel-level labels for medical images take expert time. A pathologist outlining every nucleus in a $1000\times1000$ H\&E tile, or a radiologist tracing four abdominal organs through an MR volume, is expensive. Semi-supervised segmentation tries to get close to fully supervised accuracy with only part of the images labelled. Most methods combine three ideas: pseudo-labels from a teacher network~\citep{lee2013pseudo,tarvainen2017mean}, consistency between predictions on differently perturbed views~\citep{laine2017temporal,sohn2020fixmatch,ouali2020cct,yang2023unimatch}, and, more recently, contrastive objectives that pull features of the same class together~\citep{chaitanya2020contrastive,liu2022reco,wang2021pixelcontrast}.

CSSL learns from \emph{class-wise patches}: windows cut from labelled images and from confidently pseudo-labelled ones, each restricted to the pixels of one class. A supervised contrastive loss~\citep{khosla2020supcon} with mixed hard negatives~\citep{kalantidis2020mochi} separates the class embeddings. The quality of that separation, measured by the Silhouette coefficient~\citep{rousseeuw1987silhouettes}, then sets the strength of Gaussian feature noise in a consistency loss: weak while the classes are mixed, stronger once they separate.

The main components of CSSL and the evaluation are:
\begin{itemize}[leftmargin=1.5em]
  \item \textbf{Feature-level masking.} The whole image is encoded once and its features are pooled over the class pixels of each window (\cref{sec:embed}), so the encoder sees normal images and keeps the context around small structures.
  \item \textbf{A boundary class.} For binary nucleus segmentation, nucleus pixels within two pixels of the background form a third contrastive class (\cref{sec:boundary}). Merged neighbours are a common failure in nucleus segmentation, and the errors concentrate on nucleus borders; the rim class gives these pixels their own contrastive identity.
  \item \textbf{Objective.} The consistency and contrastive losses are added to the full weak-to-strong recipe loss (\cref{sec:objective}). On MoNuSeg, a FIFO memory bank of past embeddings~\citep{wu2018instance,he2020moco} adds positives and negatives.
  \item \textbf{Evaluation protocol.} Each dataset uses the backbone, splits, labelled fraction and metric of its published baseline: Self-Loop~\citep{li2020selfloop} for MoNuSeg and DAFNet~\citep{chartsias2021dafnet} for CHAOS. The recipe alone, with every CSSL component removed, is run on the same three splits. Each final model was tested once (\cref{sec:fair}).
\end{itemize}

On MoNuSeg, CSSL improves on the recipe by \nMgainRecAbs{} nuclei F1 on average and on all three label splits, and, reading Self-Loop's F1 as nuclei F1 (\cref{sec:datasets}), exceeds the published Self-Loop result by \nMgainSLAbs{} points. On CHAOS, CSSL exceeds DAFNet by \nCgainDAFAbs{} Dice; the recipe alone is \nCrecVsDAFBAbs{} above DAFNet, and CSSL is on par with it (\nCgainRecB). In the ablations, removing the contrastive loss costs \nMDNoCLAbs{} points on MoNuSeg and \nCDNoCLAbs{} on CHAOS; the consistency term contributes less, and the memory bank lowered the test score on MoNuSeg.

\section{Related work}
\label{sec:related}

\paragraph{Semi-supervised segmentation.}
Pseudo-labelling~\citep{lee2013pseudo} and consistency regularisation~\citep{laine2017temporal,tarvainen2017mean} underlie most current methods. FixMatch~\citep{sohn2020fixmatch} combined them by training a strongly augmented view on confident pseudo-labels from a weak view. For segmentation, CutMix~\citep{yun2019cutmix} turned out to be a particularly effective strong perturbation~\citep{french2020strong}. Other methods enforce agreement between decoders~\citep{ouali2020cct} or between two networks~\citep{chen2021cps}, or make use of unreliable pixels~\citep{wang2022u2pl}. UniMatch~\citep{yang2023unimatch} adds a feature-perturbation stream and dual strong views to FixMatch and is a strong, simple recipe on both natural and medical images. Our recipe follows UniMatch's medical branch with an EMA teacher. In medical imaging, uncertainty-aware mean teachers~\citep{yu2019uamt}, disentangled representations~\citep{chartsias2019sdnet,chartsias2021dafnet} and self-supervised pseudo-label refinement~\citep{li2020selfloop} have been proposed.

\paragraph{Contrastive learning for dense prediction.}
Instance-discrimination methods~\citep{wu2018instance,he2020moco,chen2020simclr} and their supervised counterpart~\citep{khosla2020supcon} learn embeddings in which positives are pulled together and negatives pushed apart. Memory banks and momentum queues enlarge the set of negatives~\citep{wu2018instance,he2020moco}. Mixing the hardest negatives creates even harder ones~\citep{kalantidis2020mochi}. For segmentation, contrast has been applied to local and global features for medical pre-training~\citep{chaitanya2020contrastive}, to sampled pixels and regions~\citep{liu2022reco}, and to pixels across images~\citep{wang2021pixelcontrast}. CSSL contrasts class-wise \emph{patches} rather than single pixels, which keeps the number of embeddings per batch small while each one sees local context.

\paragraph{Baselines.}
Self-Loop~\citep{li2020selfloop} refines pseudo-labels with a jigsaw self-supervised task and reports results on MoNuSeg. DAFNet~\citep{chartsias2021dafnet} disentangles anatomy and modality factors for multimodal semi-supervised segmentation and reports results on CHAOS. We compare against their published numbers under their protocols, and add a recipe-only baseline trained with the same code~\citep{oliver2018realistic}.

\section{Method}
\label{sec:method}

\begin{figure}[t]
  \centering
  \includegraphics[width=\linewidth]{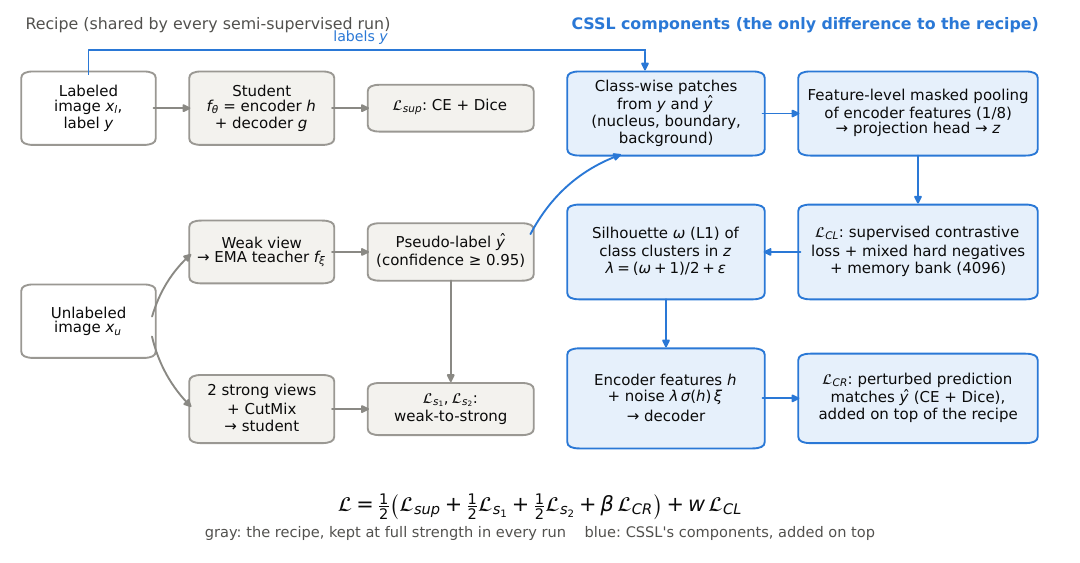}
  \caption{\textbf{CSSL on top of the recipe.} Gray: the weak-to-strong recipe, identical in every semi-supervised run of this paper. Blue: CSSL's components. Class-wise patches from labels and confident pseudo-labels are embedded by masked pooling of encoder features. A supervised contrastive loss with mixed hard negatives and a memory bank separates them. The Silhouette coefficient of the embedding clusters sets the noise level~$\lambda$ of a feature-perturbation consistency loss. The diagram shows the MoNuSeg configuration. On CHAOS, embeddings come from pixel-masked patches, no memory bank is used, and the boundary class applies to binary tasks only.}
  \label{fig:method}
\end{figure}

\subsection{Setting and recipe}
\label{sec:recipe}

We have a labelled set $\mathcal{D}_l=\{(x_i,y_i)\}$ and an unlabelled set $\mathcal{D}_u=\{u_j\}$ of 2D images with $K$ classes (background included). The network $f_\theta=g\circ h$ has an encoder $h$, which produces features $h^{(1)},\dots,h^{(L)}$ at decreasing resolution, and a decoder $g$ with skip connections. An EMA teacher $f_\xi$ with $\xi\leftarrow m_t\xi+(1-m_t)\theta$ and $m_t=\min\bigl(0.999,\tfrac{1+t}{10+t}\bigr)$~\citep{tarvainen2017mean} produces pseudo-labels and is the network we evaluate.

The supervised loss averages cross-entropy and Dice~\citep{milletari2016vnet}:
\begin{equation}
  \Lsup=\tfrac12\bigl[\CE\bigl(f_\theta(x),y\bigr)+\Dice\bigl(f_\theta(x),y\bigr)\bigr].
  \label{eq:sup}
\end{equation}
For an unlabelled image with weak view $u^w$, the teacher gives class probabilities $q=f_\xi(u^w)$, a pseudo-label $\hat y=\arg\max_k q_k$ and a confidence mask $M=\mathbb{1}[\max_k q_k\ge\delta]$, with $\delta=0.95$. Two strong views $u^{s_1},u^{s_2}$ are made with independent photometric augmentation and CutMix~\citep{yun2019cutmix} against a second unlabelled image, whose pseudo-label and mask are pasted in the same box. Each strong view is trained on its pseudo-label with a masked Dice loss, $\mathcal{L}_{s_k}=\Dice_{M}\bigl(f_\theta(u^{s_k}),\hat y\bigr)$. The \emph{recipe} objective is
\begin{equation}
  \mathcal{L}_{\mathrm{rec}}=\tfrac12\Bigl(\Lsup+\tfrac12\mathcal{L}_{s_1}+\tfrac12\mathcal{L}_{s_2}\Bigr).
  \label{eq:recipe}
\end{equation}
Training starts with $T_0=2000$ supervised iterations, after which the unlabelled losses are switched on.

\subsection{Class-wise patches}
\label{sec:patches}

For each image in the batch, we build a label map $\ell$: the ground truth $y$ for labelled images, and for unlabelled ones the pseudo-label $\hat y$ with low-confidence pixels ($M=0$) set to ``ignore''. Windows $W$ of $s\times s$ pixels ($s=64$) slide over the image with stride $r=32$. For class $k$ let $n_k(W)=\sum_{p\in W}\mathbb{1}[\ell_p=k]$ be its pixel count in the window. A window gives a patch $(W,k)$ with binary mask $m_{W,k}(p)=\mathbb{1}[\ell_p=k]$ whenever
\begin{equation}
  \begin{aligned}
  &n_k(W)\ge n_{\min} && \text{(``each'' mode), or}\\
  &k=\arg\max_{k'}n_{k'}(W)\ \text{and}\ n_k(W)\ge n_{\min} && \text{(``prominent'' mode),}
  \end{aligned}
  \label{eq:patch}
\end{equation}
with $n_{\min}=25$. Nuclei are rarely the most frequent class in a $64\times64$ window, so MoNuSeg uses the ``each'' mode and treats background as a class. CHAOS uses the ``prominent'' mode over the four organs. At most $816$ patches per batch are kept, drawn round-robin over classes so that rare classes survive.

\subsection{Patch embeddings: pixel-level versus feature-level masking}
\label{sec:embed}

\paragraph{Pixel-level masking.} The other classes' pixels of each patch are set to zero, the patch is encoded and the bottleneck features are pooled over the (downsampled) mask:
\begin{equation}
  z_{W,k}=\nrm\Bigl(\phi\Bigl(\frac{\sum_q \bar m_{W,k}(q)\,h^{(L)}_q\bigl(x_W\odot m_{W,k}\bigr)}{\sum_q \bar m_{W,k}(q)}\Bigr)\Bigr),
  \label{eq:pixelmask}
\end{equation}
where $\phi$ is a two-layer projection head into $\mathbb{R}^{256}$, $\nrm(v)=v/\lVert v\rVert_2$, $q$ indexes feature cells and $\bar m$ is the mask downsampled by area averaging, i.e., the fraction of each cell's pixels that belong to class $k$.

\begin{figure}[t]
  \centering
  \includegraphics[width=\linewidth]{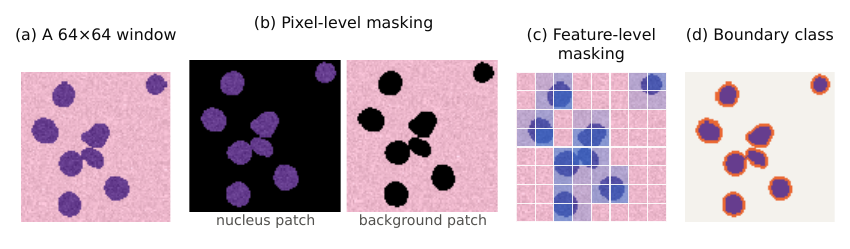}
  \caption{\textbf{What the contrastive branch sees} (synthetic illustration). (a) A $64\times64$ window. (b) Pixel-level masking: a nucleus patch and a background patch with the pixels of the other class set to zero. (c) Feature-level masking encodes the unmasked image and pools encoder features over the cells covered by the class. The cell shading shows the pooling weight $\bar m$. (d) The boundary class: nucleus pixels within $w=2$ pixels of the background (orange) form a separate contrastive class.}
  \label{fig:embedding}
\end{figure}

\paragraph{Feature-level masking.} The full, unmasked image is encoded once and the features of level $\ell$ are pooled over the window's class pixels:
\begin{equation}
  z_{W,k}=\nrm\Bigl(\phi\Bigl(\frac{\sum_{q\in W} \bar m_{W,k}(q)\,h^{(\ell)}_q(x)}{\sum_{q\in W} \bar m_{W,k}(q)}\Bigr)\Bigr).
  \label{eq:featmask}
\end{equation}
With feature-level masking (\cref{fig:embedding}), the encoder only sees real images, and the embeddings use the same features the decoder segments from. Context outside the class mask (a nucleus's surroundings) informs the embedding through the receptive field, and the mask only weights the pooling. MoNuSeg uses feature-level masking at $1/8$ resolution; CHAOS uses pixel-level masking. Batch-norm running statistics are frozen during embedding passes.

\subsection{Boundary class for touching objects}
\label{sec:boundary}

In nucleus segmentation the costly errors are merged neighbours and inaccurate borders, both of which happen at rim pixels. For binary tasks we relabel the contrastive label map, leaving the segmentation targets unchanged:
\begin{equation}
  \ell'_p=\begin{cases}K & \text{if } \ell_p=1 \text{ and } \min_{p':\,\ell_{p'}=0}\lVert p-p'\rVert_\infty\le w,\\ \ell_p & \text{otherwise,}\end{cases}
  \label{eq:boundary}
\end{equation}
with $w=2$. This is computed as a $(2w{+}1)\times(2w{+}1)$ max-pool of the background indicator. Patches are then sampled from $\ell'$ with three classes (background, nucleus interior, rim). Pseudo-labelled pixels are relabelled only if they are confident.

\subsection{Contrastive loss with hard negatives and a memory bank}
\label{sec:cl}

Let $\{(z_i,k_i)\}_{i=1}^{N}$ be the patch embeddings of the batch and their labels, and let $\mathcal{Q}$ be a FIFO memory of the $4096$ most recent embeddings with labels, stored without gradient. For anchor $i$, the positives are $P(i)=\{j\neq i: k_j=k_i\}$ and the negatives $N(i)=\{j: k_j\neq k_i\}$, both taken over the batch and $\mathcal{Q}$. Following \citet{kalantidis2020mochi}, we synthesise $H(i)=\min(|N(i)|,256)$ harder negatives per anchor by mixing the anchor with random negatives:
\begin{equation}
  \tilde z_{i,t}=\sg\bigl[\nrm\bigl(\alpha_t z_i+(1-\alpha_t)z_{n_t}\bigr)\bigr],\qquad n_t\sim\mathcal{U}\bigl(N(i)\bigr),\quad \alpha_t\sim\mathcal{U}(0,0.4),
  \label{eq:hard}
\end{equation}
where $\sg$ stops the gradient. This gives a $1{:}1$ ratio of hard to ordinary negatives, up to the cap. With similarities $s_{ij}=z_i^\top z_j/\tau$ and $\tau=0.1$, the loss contrasts each positive with all negatives:
\begin{equation}
  \Lcl=\frac{1}{|\mathcal{A}|}\sum_{i\in\mathcal{A}}\frac{-1}{|P(i)|}\sum_{p\in P(i)}\log\frac{e^{s_{ip}}}{e^{s_{ip}}+\sum_{n\in N(i)}e^{s_{in}}+\sum_{t=1}^{H(i)}e^{z_i^\top\tilde z_{i,t}/\tau}},
  \label{eq:cl}
\end{equation}
where $\mathcal{A}$ is the set of batch anchors with at least one positive and one negative. Only current embeddings act as anchors, and the memory supplies extra positives and negatives. If a batch contains a single class, $256$ samples of a fixed random Gaussian are added as negatives so that the loss stays defined.

\subsection{Silhouette-adaptive consistency}
\label{sec:cr}

The Silhouette coefficient~\citep{rousseeuw1987silhouettes} summarises cluster separation in $[-1,1]$. With the Manhattan distance $d$ on the batch embeddings (at most 1024, subsampled), and for each embedding $i$ with class $k_i$,
\begin{equation}
  a_i=\frac{\sum_{j:k_j=k_i,\,j\ne i}d(z_i,z_j)}{|\{j:k_j=k_i\}|-1},\qquad
  b_i=\min_{k\neq k_i}\frac{\sum_{j:k_j=k}d(z_i,z_j)}{|\{j:k_j=k\}|},\qquad
  \omega=\frac1N\sum_i\frac{b_i-a_i}{\max(a_i,b_i)}.
  \label{eq:sil}
\end{equation}
A single batch is noisy, so we smooth it, $\bar\omega_t=\mu\bar\omega_{t-1}+(1-\mu)\omega_t$ with $\mu=0.99$ and $\bar\omega_0=0$, and map it to a noise level
\begin{equation}
  \lambda_t=\frac{\bar\omega_t+1}{2}+\varepsilon,\qquad \varepsilon=0.05.
  \label{eq:lambda}
\end{equation}
Every encoder feature map of the weak view, skip connections included, is perturbed with noise scaled to the map's own spread,
\begin{equation}
  \tilde h^{(l)}=h^{(l)}+\sqrt{\lambda_t}\;\sigma\bigl(h^{(l)}\bigr)\,\xi,\qquad \xi\sim\mathcal{N}(0,I),\quad l=1,\dots,L,
  \label{eq:noise}
\end{equation}
where $\sigma(\cdot)$ is the per-sample standard deviation of the map (no gradient). The decoder's prediction from the perturbed features must match the teacher's pseudo-label on confident pixels:
\begin{equation}
  \Lcr=\tfrac12\bigl[\CE_M\bigl(g(\tilde h),\hat y\bigr)+\Dice_M\bigl(g(\tilde h),\hat y\bigr)\bigr].
  \label{eq:cr}
\end{equation}
While the classes are mixed in embedding space ($\bar\omega\approx0$), the perturbation stays moderate. As the contrastive loss separates them, it grows, so the decoder is asked to be robust to larger feature changes only once the features support it.

\subsection{Objective}
\label{sec:objective}

CSSL adds its two losses to the recipe objective:
\begin{equation}
  \mathcal{L}=\tfrac12\Bigl(\Lsup+\tfrac12\mathcal{L}_{s_1}+\tfrac12\mathcal{L}_{s_2}+\beta\,\Lcr\Bigr)+w(t)\,\Lcl,\qquad
  w(t)=w_0\exp\Bigl(-5\bigl(1-\min(1,\tfrac{t-T_0}{T_r})\bigr)^2\Bigr),
  \label{eq:total}
\end{equation}
with $\beta=0.5$, $w_0=0.1$ and a ramp of $T_r=1000$ iterations after the supervised warm-up~\citep{laine2017temporal}. Setting $\beta=0$ and $w_0=0$ recovers the recipe~\eqref{eq:recipe} exactly, which makes it the control in all our comparisons. \Cref{alg:cssl} summarises one training step.

\begin{algorithm}[t]
\caption{One CSSL training step (after the supervised warm-up)}
\label{alg:cssl}
\small
\begin{algorithmic}[1]
\State Sample labelled $(x,y)$ and unlabelled $u,u'$; make the weak view $u^w$ and strong views $u^{s_1},u^{s_2}$ (CutMix with $u'$)
\State $q\gets f_\xi(u^w)$; \; $\hat y\gets\arg\max q$; \; $M\gets\mathbb{1}[\max q\ge\delta]$ \Comment{teacher pseudo-labels}
\State $\mathcal{L}_{\mathrm{rec}}\gets$ \cref{eq:recipe} \Comment{recipe, unchanged}
\State $\Lcr\gets$ \cref{eq:cr} with features perturbed by \cref{eq:noise} at the current $\lambda_t$
\State $\ell\gets[y;\hat y\text{ with }M{=}0\text{ ignored}]$; optionally relabel rims (\cref{eq:boundary}); sample patches (\cref{eq:patch})
\State $z\gets$ feature-level embeddings (\cref{eq:featmask}) with BN statistics frozen
\State $\Lcl\gets$ \cref{eq:cl} with hard negatives (\cref{eq:hard}) and memory $\mathcal{Q}$; push $\sg[z]$ into $\mathcal{Q}$
\State $\omega_t\gets$ \cref{eq:sil}; \; $\bar\omega_t\gets\mu\bar\omega_{t-1}+(1-\mu)\omega_t$; \; $\lambda_{t+1}\gets(\bar\omega_t+1)/2+\varepsilon$
\State Update $\theta$ on $\tfrac12(\Lsup+\tfrac12\mathcal{L}_{s_1}+\tfrac12\mathcal{L}_{s_2}+\beta\Lcr)+w(t)\Lcl$; \; $\xi\gets m_t\xi+(1-m_t)\theta$
\end{algorithmic}
\end{algorithm}

\section{Experimental protocol}
\label{sec:protocol}

\subsection{Datasets, baselines and metrics}
\label{sec:datasets}

\paragraph{MoNuSeg, the protocol of Self-Loop.} MoNuSeg~\citep{kumar2017monuseg,kumar2020monuseg} has 30 training and 14 test H\&E images ($1000\times1000$, $40\times$) with nucleus annotations. Following Self-Loop~\citep{li2020selfloop}, 50\% of the training images (15) are labelled, the other 15 are unlabelled, the network is Self-Loop's ResUNet-18~\citep{he2016resnet,ronneberger2015unet}, randomly initialised since Self-Loop mentions no pre-training (14.4M parameters in our implementation), whole test images are segmented, and the final weights are tested. Self-Loop does not publish its labelled subset, so we draw three random subsets (seeds 0, 1, 2) and report all three.

The metric needs care. Self-Loop reports an ``F1 score'', described as the unweighted average of foreground and background F1. Its own supervised baselines (75.87 at 50\% labels, 79.30 at 100\%) are close to our supervised \emph{nuclei-only} F1 under the same protocol (\nMsupFifty{} and \nMsupHundred{}), whereas the foreground--background average or balanced accuracy would be 5--9 points higher (\cref{fig:calibration}). We therefore compare on nuclei F1, i.e., the per-image Dice of the nucleus class averaged over the 14 test images. Of the three readings, this is also the strictest for us.

\paragraph{CHAOS, the protocol of DAFNet.} CHAOS~\citep{kavur2021chaos} contains abdominal MR scans of 20 subjects with liver, kidney and spleen labels. DAFNet~\citep{chartsias2021dafnet} uses T1-DUAL (the out-phase images, as in its code) as a fully labelled source modality and T2-SPIR as the partially labelled target. We use DAFNet's three train/validation/test splits (14/3/3 subjects) and its labelled subjects at 25\% (4 T2 subjects, chosen by DAFNet's own seeded sampling), with T1 labels for all training subjects, as in DAFNet. The other 10 training subjects' T2 scans are unlabelled. The network is a U-Net with the widths of DAFNet's U-Net (64 to 1024 channels, four down-samplings; 29.0M parameters in our implementation), with a single T2 slice of $192\times192$ as input. We use DAFNet's test formula: per slice, the Dice of the four organ channels pooled, averaged over slices and subjects. Our predicted masks are the argmax of the softmax output, whereas DAFNet's code rounds each organ probability separately; the two differ only on pixels whose most likely class has probability at most 0.5.\ifhaveround{} With DAFNet's rounding, the mean scores of the same checkpoints change by at most \nCroundMaxDiff{} points (\cref{app:secondary}).\fi{} Following DAFNet, the checkpoint with the best validation Dice (argmax masks) is tested.

\paragraph{Reported baselines.} Baseline numbers are taken from the respective papers: Self-Loop's Table~2 at 50\% labels and DAFNet's table at 25\%. Our own runs cover a supervised model (labelled data only), the recipe alone, and CSSL, all with identical code, data, network and schedule.

\subsection{Implementation}
All runs use SGD (learning rate 0.01, momentum 0.9, weight decay $10^{-4}$, polynomial decay with power 0.9), bf16 autocast, and the losses and teacher described in \cref{sec:method}. MoNuSeg uses batches of 8 labelled and 8 unlabelled $256\times256$ crops, 30k iterations, random-scale crops and H\&E stain augmentation. CHAOS uses 12+12 slices, 16k iterations, affine augmentation without flips (to keep left and right organs apart), and validation every 1000 iterations. Hyperparameters are listed in \cref{app:hparams}.

\subsection{Evaluation protocol}
\label{sec:fair}
\begin{enumerate}[leftmargin=1.5em]
  \item \textbf{Baseline protocols}: backbone, input, labelled fraction, splits (DAFNet's own on CHAOS), metric and tested checkpoint follow the baseline. Our supervised MoNuSeg model scores \nMsupFifty, close to Self-Loop's supervised model (\nSLsupFifty).
  \item \textbf{Recipe-only baseline} on the same splits, with $\beta=w_0=0$ in \cref{eq:total}.
  \item \textbf{Comparison rule}: a method is better than another if its mean over the three splits is higher and it is higher on at least two of the three. Standard deviations are population SDs over the three runs.
  \item No test-time augmentation or post-processing; single models.
\end{enumerate}

\section{Results}
\label{sec:results}

\begin{figure}[t]
  \centering
  \includegraphics[width=\linewidth]{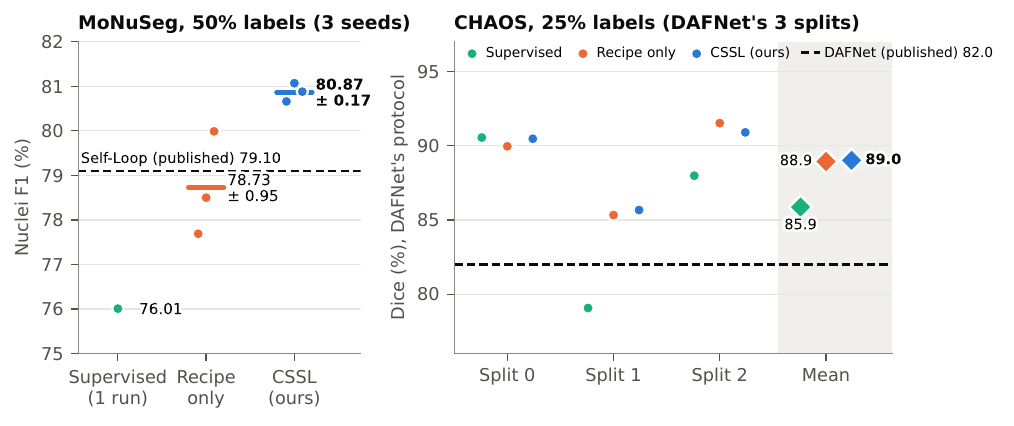}
  \caption{\textbf{Main results.} Dots are individual label splits (MoNuSeg seeds; DAFNet's CHAOS splits). Horizontal ticks (MoNuSeg) and diamonds (CHAOS) are means. Dashed lines are the published baselines. CHAOS uses the best-validation checkpoint (DAFNet's rule).}
  \label{fig:results}
\end{figure}

\subsection{MoNuSeg}

\begin{table}[t]
  \centering
  \caption{\textbf{MoNuSeg, 50\% labels} (15 of 30 training images), nuclei F1 (\%) on the 14 test images, final weights. Published numbers are single runs on an unspecified labelled subset. Ours are three random labelled subsets with the same ResUNet-18.}
  \label{tab:monuseg}
  \small
  \begin{tabular}{llccccc}
    \toprule
    & Method & Seed 0 & Seed 1 & Seed 2 & Mean\pm SD & vs.\ recipe \\
    \midrule
    \multirow{5}{*}{\rotatebox{90}}
    & Supervised (Self-Loop's report) & \multicolumn{3}{c}{--} & \nSLsupFifty & \\
    & Softmax confidence & \multicolumn{3}{c}{--} & \nSLtSoftmax & \\
    & Ensemble & \multicolumn{3}{c}{--} & \nSLtEnsemble & \\
    & MC Dropout & \multicolumn{3}{c}{--} & \nSLtMcDropout & \\
    & Self-Loop~\citep{li2020selfloop} & \multicolumn{3}{c}{--} & \nSL & \\
    \midrule
    \multirow{3}{*}{\rotatebox{90}}
    & Supervised (labelled only) & \nMsupFifty & -- & -- & \nMsupFifty & \\
    & Recipe only & \nMrecsa & \nMrecsb & \nMrecsc & \nMrecMean\pm\nMrecSd & \\
    \rowcolor{rowhl} & \ours{CSSL (ours)} & \ours{\nMcsslsa} & \ours{\nMcsslsb} & \ours{\nMcsslsc} & \ours{\nMcsslMean\pm\nMcsslSd} & \ours{\nMgainRec{}, \nMwinsRec/3} \\
    \midrule
    & \textit{Reference: supervised, 100\% labels} & \textit{\nMsupHundred} & -- & -- & \textit{\nMsupHundred} & \\
    \bottomrule
  \end{tabular}
\end{table}

\Cref{tab:monuseg} and \cref{fig:results} (left) give the MoNuSeg results. Our supervised model reproduces Self-Loop's supervised number (\nMsupFifty{} vs.\ \nSLsupFifty). The recipe alone reaches \nMrecMean\pm\nMrecSd, slightly below Self-Loop's \nSL{} on average, with a spread across label subsets (\nMrecsa{} to \nMrecsc) larger than most gaps in Self-Loop's table. CSSL reaches \nMcsslMean\pm\nMcsslSd: \nMgainRecAbs{} above the recipe, higher on all three seeds (per-seed gains \nMgainReca, \nMgainRecb, \nMgainRecc), and, with Self-Loop's F1 read as nuclei F1, \nMgainSLAbs{} above Self-Loop's published number, above it on all three seeds. It also has a much smaller spread than the recipe. For reference, a fully supervised model trained on all 30 images reaches \nMsupHundred{} under the same protocol, which is \nMgainSupHundredAbs{} below CSSL at half the labels (one seed). The boundary class adds the most in the ablation (\cref{sec:ablation}).

\begin{figure}[t]
  \centering
  \includegraphics[width=\linewidth]{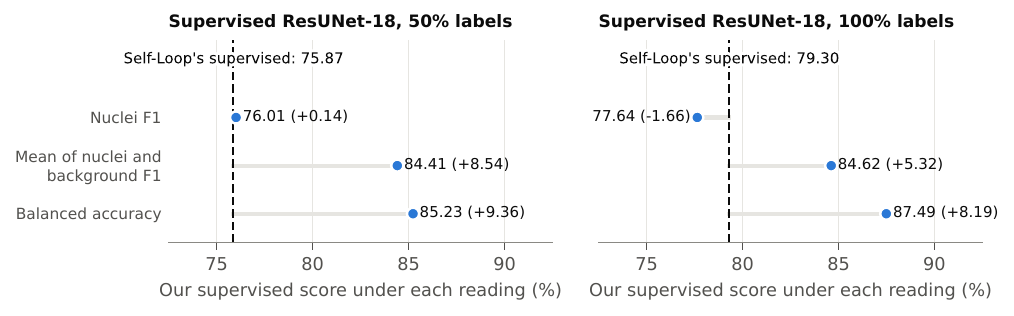}
  \caption{\textbf{Calibrating the metric.} Our supervised ResUNet-18, in Self-Loop's protocol, scored three ways. Only nuclei F1 (per-image nucleus Dice) matches the supervised numbers Self-Loop reports, so we compare on it.}
  \label{fig:calibration}
\end{figure}

\paragraph{Which F1 does Self-Loop report?} \Cref{fig:calibration} scores our supervised models under the three readings of Self-Loop's metric. Only nuclei F1 lands near Self-Loop's supervised numbers, at 50\% (\nMsupGapSL) and at 100\% labels (\nMsupHGapSL). The other readings are 5--9 points higher. Under the other two readings even our supervised model would exceed 79.10 (\nMsupFiftyFgbg{} and \nMsupFiftyBacc), so nuclei F1 is the most conservative comparison. Self-Loop's evaluation code is not public, so this calibration supports the nuclei-F1 reading without proving it; the comparison with the recipe uses our code on both sides and does not depend on it. We report the secondary metrics in \cref{app:secondary}.

\subsection{CHAOS}

\begin{table}[t]
  \centering
  \caption{\textbf{CHAOS T2, 25\% labels} (4 labelled T2 subjects plus T1 labels, DAFNet's splits), DAFNet's Dice (\%) on each split's 3 test subjects. Best-validation checkpoint (DAFNet's rule). Published numbers are DAFNet's Table II (averages over its three splits).}
  \label{tab:chaos}
  \small\setlength{\tabcolsep}{4.5pt}
  \begin{tabular}{llcccc}
    \toprule
    & Method & Split 0 & Split 1 & Split 2 & Mean\pm SD \\
    \midrule
    \multirow{5}{*}{\rotatebox{90}}
    & U-Net (single / multi-modal) & \multicolumn{3}{c}{--} & \nDtUNetSingle{} / \nDtUNetMulti \\
    & SDNet (single / multi-modal)~\citep{chartsias2019sdnet} & \multicolumn{3}{c}{--} & \nDtSdnetSingle{} / \nDtSdnetMulti \\
    & Dual-stream / Translation & \multicolumn{3}{c}{--} & \nDtDualstream{} / \nDtTranslation \\
    & DADR & \multicolumn{3}{c}{--} & \nDtDadr \\
    & DAFNet~\citep{chartsias2021dafnet} & \multicolumn{3}{c}{--} & \nDAF \\
    \midrule
    \multirow{3}{*}{\rotatebox{90}}
    & Supervised (labelled only) & \nCsupBa & \nCsupBb & \nCsupBc & \nCsupBMean\pm\nCsupBSd \\
    & Recipe only & \nCrecBa & \nCrecBb & \nCrecBc & \nCrecBMean\pm\nCrecBSd \\
    \rowcolor{rowhl} & \ours{CSSL (ours)} & \ours{\nCcsslBa} & \ours{\nCcsslBb} & \ours{\nCcsslBc} & \ours{\nCcsslBMean\pm\nCcsslBSd} \\
    \bottomrule
  \end{tabular}
\end{table}

On CHAOS (\cref{tab:chaos}), CSSL reaches \nCcsslBMean{} Dice, \nCgainDAFAbs{} above DAFNet and above it on every split. DAFNet reports \nDtUNetMulti{} for a U-Net trained on both modalities. Our supervised U-Net, trained on the same labelled subjects, reaches \nCsupBMean, and the recipe alone reaches \nCrecBMean. Against the recipe, CSSL is \nCgainRecB{} (higher on \nCwinsRecB{} of 3 splits). The difference is smaller than the variation between splits, so CSSL and the recipe perform on par on CHAOS. On split 1, where the supervised model is weakest (\nCsupBb), the semi-supervised methods gain about six points, and CSSL is marginally the best of them.

\section{Ablations}
\label{sec:ablation}

\begin{table}[t]
  \centering
  \caption{\textbf{Ablation of the final CSSL}, test set, one label split each (MoNuSeg seed 0, CHAOS split 1). $\Delta$ is the full model minus the ablated one, so a positive value means the component helps. ``w/o CL'' also freezes $\lambda$ at its initial value, since the Silhouette coefficient comes from the contrastive embeddings. ``w/o Silhouette'' fixes $\lambda$ at the run's average.}
  \label{tab:ablation}
  \small
  \begin{tabular}{lcccc}
    \toprule
    & \multicolumn{2}{c}{MoNuSeg (nuclei F1)} & \multicolumn{2}{c}{CHAOS (Dice)} \\
    \cmidrule(lr){2-3}\cmidrule(lr){4-5}
    Variant & Score & $\Delta$ & Score & $\Delta$ \\
    \midrule
    \rowcolor{rowhl} \ours{Full CSSL} & \ours{\nMAFull} & & \ours{\nCAFull} & \\
    w/o contrastive loss (CL) & \nMANoCL & \nMDNoCL & \nCANoCL & \nCDNoCL \\
    w/o consistency loss (CR) & \nMANoCR & \nMDNoCR & \nCANoCR & \nCDNoCR \\
    w/o hard negatives (HNM) & \nMANoHNM & \nMDNoHNM & \nCANoHNM & \nCDNoHNM \\
    w/o Silhouette feedback (fixed $\lambda$) & \nMANoSil & \nMDNoSil & \nCANoSil & \nCDNoSil \\
    w/o boundary class & \nMANoBnd & \nMDNoBnd & \multicolumn{2}{c}{\textit{not used}} \\
    w/o feature-level masking & \nMANoFeat & \nMDNoFeat & \multicolumn{2}{c}{\textit{not used}} \\
    w/o memory bank & \nMANoMem & \nMDNoMem & \multicolumn{2}{c}{\textit{not used}} \\
    \midrule
    Recipe only (all of CSSL removed) & \nMARec & \nMDRec & \nCARec & \nCDRec \\
    \bottomrule
  \end{tabular}
\end{table}

\begin{figure}[t]
  \centering
  \includegraphics[width=\linewidth]{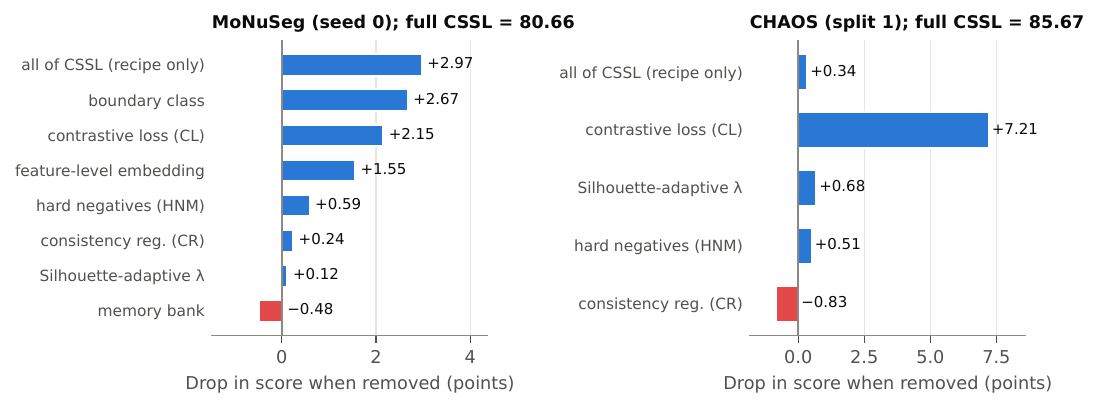}
  \caption{\textbf{Ablation effects} (full model minus ablated model, test set, one split each). Blue: the component helps. Red: the score is higher without it.}
  \label{fig:ablation}
\end{figure}

\Cref{tab:ablation} and \cref{fig:ablation} remove one component at a time from the final configuration of each dataset. Each row is a single run on a single split; for reference, the recipe varies by 2.3 points across MoNuSeg seeds.

\paragraph{The contrastive branch is the core of the method.} Removing the contrastive loss costs \nMDNoCLAbs{} points on MoNuSeg and \nCDNoCLAbs{} on CHAOS. On MoNuSeg the three components that only act through the contrastive loss are also the largest contributors: the boundary class (\nMDNoBnd), feature-level masking (\nMDNoFeat) and hard negatives (\nMDNoHNM). On CHAOS, hard negatives add \nCDNoHNMAbs. On MoNuSeg, removing the boundary class alone costs more (\nMDNoBndAbs) than removing the contrastive loss altogether (\nMDNoCLAbs): with only nucleus and background patches, the contrastive term did not help in this configuration.

\paragraph{Consistency regularisation is weak, and needs the contrastive loss.} CR adds \nMDNoCRAbs{} on MoNuSeg. On CHAOS split 1 the model scores higher without it (\nCDNoCR). Removing CL while keeping CR is worse than the recipe on CHAOS (\nCANoCL{} vs.\ \nCARec). Perturbing features at a constant noise level, without the contrastive term shaping them, hurts there.

\paragraph{Silhouette feedback.} With $\lambda$ fixed at its average value, the score drops by \nMDNoSilAbs{} (MoNuSeg) and \nCDNoSilAbs{} (CHAOS).

\paragraph{Memory bank.} On seed 0, the model without the memory bank scores \nMANoMem, \nMDNoMemAbs{} above the full model.

\IfFileExists{figures/fig_qual_monuseg.pdf}{\section{Qualitative results}
\label{sec:qual}

\Cref{fig:qual-monuseg,fig:qual-chaos} show predictions of the evaluated networks (EMA teachers with the reported weights). For MoNuSeg we show the test images with the largest, median and smallest per-image gain of CSSL over the recipe, and in each image the $256\times256$ window with the most annotated nucleus pixels. For CHAOS we show, in each split, the test slice with the largest annotated organ area. On the MoNuSeg image with the largest gain, the supervised model and the recipe mark large background regions as nuclei (per-image nuclei F1 33.9 and 41.3), while CSSL does not (79.0). On the median and smallest-gain images the predictions of the three models are close (CSSL and the recipe differ by at most 0.6 F1). On the CHAOS split 1 slice, CSSL recovers more of the liver and part of the spleen than the recipe (slice Dice 43.4 vs.\ 27.1); on splits 0 and 2 the predictions of all three models are close.

\begin{figure}[t]
  \centering
  \includegraphics[width=0.98\linewidth]{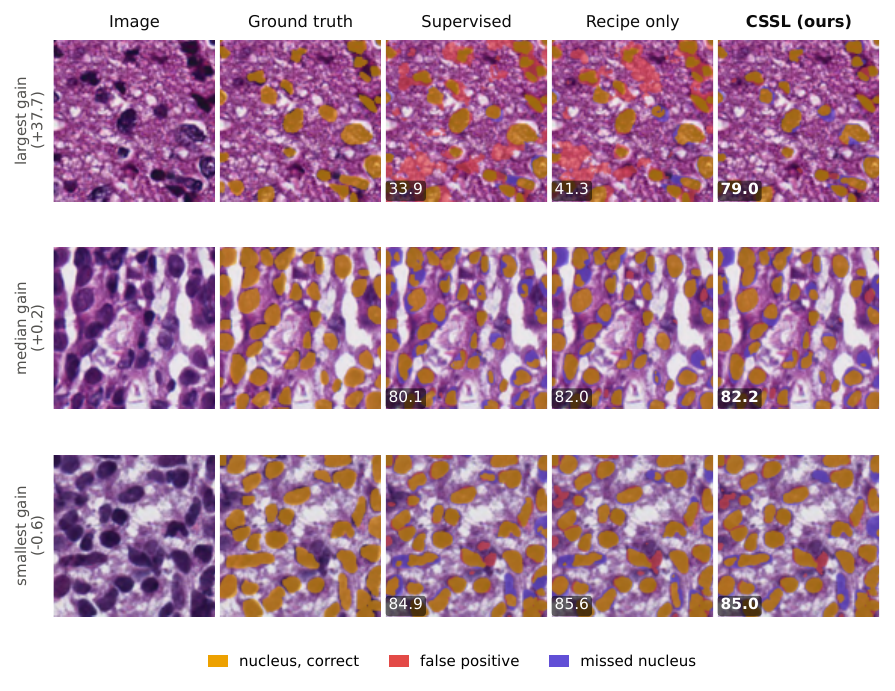}
  \caption{\textbf{MoNuSeg test windows} (seed 0). Yellow: correctly segmented nucleus pixels. Red: false positives. Violet: missed nucleus pixels. Numbers are the nuclei F1 of the whole test image. Rows: the images with the largest, median and smallest gain of CSSL over the recipe (the gain in parentheses).}
  \label{fig:qual-monuseg}
\end{figure}

\begin{figure}[t]
  \centering
  \includegraphics[width=0.98\linewidth]{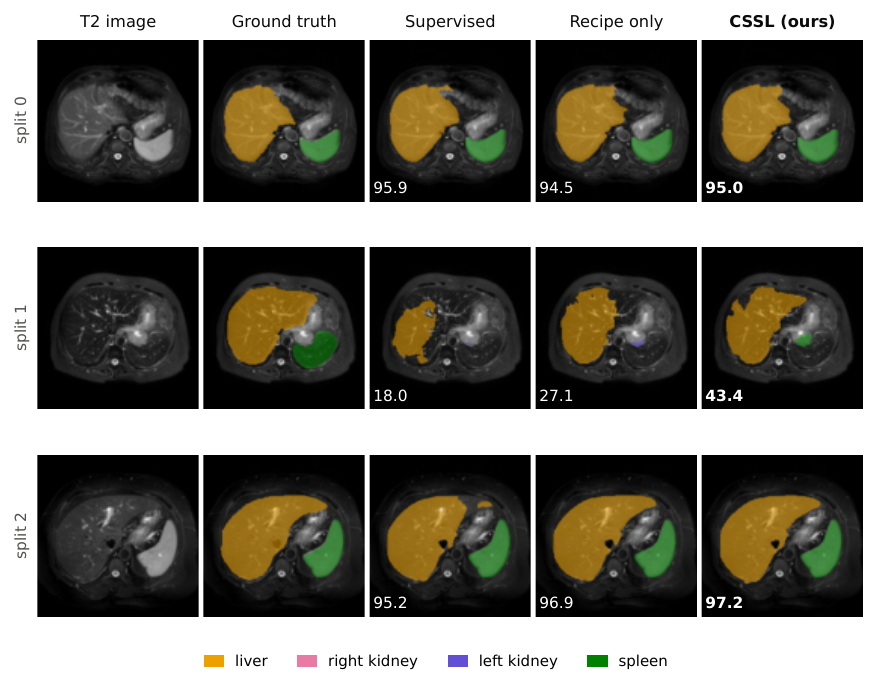}
  \caption{\textbf{CHAOS T2 test slices}, one per DAFNet split (the slice with the largest annotated organ area), best-validation checkpoints. Numbers are the mean organ Dice on the shown slice.}
  \label{fig:qual-chaos}
\end{figure}
}{}

\section{Discussion}
\label{sec:discussion}

\paragraph{Contribution of CSSL.} On MoNuSeg, the recipe alone is below Self-Loop on average, our supervised model matches Self-Loop's, and CSSL adds \nMgainRecAbs{} on average, with a gain on every label subset. On CHAOS, the recipe alone exceeds DAFNet by about seven points, and CSSL performs on par with it.

\paragraph{MoNuSeg versus CHAOS.} Nuclei are small, numerous and often touch, so patch-level contrast with a boundary class supplies many informative, local positives and negatives per image. CHAOS organs are few and large, each slice gives only a handful of windows per organ, and the recipe alone already performs well.

\section{Conclusion}
\label{sec:conclusion}

We presented CSSL, a contrastive semi-supervised segmentation method with class-wise patches, feature-level masked pooling, a boundary class, mixed hard negatives, a memory bank and Silhouette-adaptive consistency, added to a weak-to-strong pseudo-labelling recipe. Under the protocols of the published baselines, CSSL reaches \nMcsslMean{} nuclei F1 on MoNuSeg with 50\% labels (Self-Loop: \nSL; recipe: \nMrecMean) and \nCcsslBMean{} Dice on CHAOS T2 with 25\% labels (DAFNet: \nDAF; recipe: \nCrecBMean). The contrastive branch accounts for most of the improvement.

\bibliography{references}
\bibliographystyle{plainnat}

\appendix
\section{Hyperparameters}
\label{app:hparams}

\begin{table}[H]
  \centering
  \caption{Hyperparameters. Where the two datasets differ, MoNuSeg / CHAOS.}
  \small
  \footnotesize
  \begin{tabular}{ll}
    \toprule
    Network & ResUNet-18, random init (14.4M) / U-Net 64--1024, random init (29.0M) \\
    Input & RGB $256\times256$ crops / one T2 slice, $192\times192$, intensities in $[-1,1]$ \\
    Batch & 8 labelled + 8 unlabelled / 12 + 12 (half T2, half T1 among labelled) \\
    Iterations & 30{,}000 / 16{,}000, of which 2{,}000 supervised warm-up \\
    Optimiser & SGD, lr 0.01, momentum 0.9, wd $10^{-4}$, poly decay (0.9) \\
    Teacher & EMA, decay $\min(0.999,\frac{1+t}{10+t})$; confidence threshold $\delta=0.95$ \\
    Strong views & 2, CutMix ($p=0.5$); H\&E stain, colour jitter, blur / intensity augmentation \\
    Patches & $64\times64$, stride 32, $n_{\min}=25$, $\le816$ per batch; ``each'' + background / ``prominent'', organs \\
    Embedding & feature-level at $1/8$ resolution / pixel-level masking at the bottleneck; 256-d, 2-layer head \\
    Boundary class & $w=2$ px / -- \\
    Contrastive & $\tau=0.1$, hard negatives $\alpha\sim\mathcal{U}(0,0.4)$, 1:1, $\le256$ per anchor, $w_0=0.1$, ramp 1000 it. \\
    Memory bank & 4096 embeddings / -- \\
    Consistency & $\beta=0.5$, $\varepsilon=0.05$, Silhouette EMA $\mu=0.99$, noise relative to feature SD \\
    Tested weights & final / best validation Dice (validation every 1000 iterations) \\
    \bottomrule
  \end{tabular}
\end{table}

\section{Secondary metrics}
\label{app:secondary}

For MoNuSeg we also give, for seed 0, the foreground--background F1 average and balanced accuracy (both per image). Supervised (50\%): \nMsupFiftyFgbg{} / \nMsupFiftyBacc. Supervised (100\%): \nMsupHundredFgbg{} / \nMsupHundredBacc.\ifhavesecondary{} CSSL: \nMcsslFgbg{} / \nMcsslBacc{} (pooled-pixel nuclei F1: \nMcsslPooled).\fi{}

\ifhaveround{}For CHAOS, \cref{tab:chaos-round} scores the checkpoints of \cref{tab:chaos} with DAFNet's rounding of each organ probability instead of the argmax.

\begin{table}[h]
  \centering\footnotesize
  \caption{\textbf{CHAOS T2, 25\% labels, with DAFNet's rounding.} Same checkpoints as \cref{tab:chaos}; each organ mask is the organ probability rounded at 0.5, as in DAFNet's code. Last column: difference from the argmax score.}
  \label{tab:chaos-round}
  \begin{tabular}{lccccc}
    \toprule
    & Split 0 & Split 1 & Split 2 & Mean $\pm$ SD & vs.\ argmax \\
    \midrule
    Supervised & \nCsupRa & \nCsupRb & \nCsupRc & \nCsupRMean{} $\pm$ \nCsupRSd & \nCsupRDiff \\
    Recipe only & \nCrecRa & \nCrecRb & \nCrecRc & \nCrecRMean{} $\pm$ \nCrecRSd & \nCrecRDiff \\
    CSSL & \nCcsslRa & \nCcsslRb & \nCcsslRc & \nCcsslRMean{} $\pm$ \nCcsslRSd & \nCcsslRDiff \\
    \bottomrule
  \end{tabular}
\end{table}
\fi{}

\end{document}